\def\year{2019}\relax
\documentclass[letterpaper]{article} 
\usepackage{aaai19}
\usepackage{times}  
\usepackage{helvet}  
\usepackage{courier}  
\usepackage{url}  
\usepackage{graphicx}  
\usepackage{subcaption}
\usepackage{multirow}

\usepackage{bm}

\begin{document}
%
\title{Merging Datasets Through Deep learning}

\author{Kavitha Srinivas \\ IBM Research
\And Abraham Gale \\ Yeshiva University
\And Julian Dolby \\ IBM Research}

\maketitle
\begin{abstract}
Merging datasets is a key operation for data analytics.  A frequent
requirement for merging is joining across columns that have
different surface forms for the same entity (e.g., the name of a
person might be represented as \textit{Douglas Adams} or
\textit{Adams, Douglas}).  Similarly,
ontology alignment can require recognizing distinct surface forms of
the same entity, especially when ontologies are independently
developed.  However, data management systems are currently limited
to performing merges based on string equality, or at best using
string similarity.  We propose an approach to performing merges
based on deep learning models.  Our approach depends on (a) creating
a deep learning model that maps surface forms of an entity into a
set of vectors such that alternate forms for the same entity are
closest in vector space, (b) indexing these vectors using a nearest
neighbors algorithm to find the forms that can be potentially joined
together.  To build these models, we had to adapt techniques from
metric learning due to the characteristics of the data; specifically we describe 
novel sample selection techniques and loss functions that work for this problem.  
To evaluate our approach, we used Wikidata as ground truth
and built models from datasets with approximately 1.1M people's names
(200K identities) and 130K company names (70K identities).  We developed models that 
allow for joins with precision@1 of .75-.81 and
recall of .74-.81.  We make the models available for aligning people or companies across multiple datasets.  
\end{abstract}

\section{Introduction}

Merging datasets is a key operation for data analytics.  A frequent
requirement for merging is joining across columns that have
different surface forms for the same entity.  For instance, the name of a person might be represented as \textit{Douglas Adams}, \textit{Douglas Noel Adams}, \textit{D. Adams} or \textit{Adams, Douglas}.  Similarly, ontology alignment can require recognizing distinct surface forms of
the same entity, especially when ontologies are independently
developed.  This problem occurs for many entity types such as people's names, company names, addresses, product descriptions, conference venues, or even people's faces.  Data management systems have however, largely focussed solely on equi-joins, where string or numeric equality determines which rows should be joined, because such joins are efficient.

We propose a different approach to joining different surface representations of the same entity, inspired by recent advances in deep learning.  Our approach depends on (a) mapping surface forms into sets of vectors such that forms for the same entity are closest in vector space, (b) indexing these vectors to find the forms that can be potentially joined together.  The approach is general, in the sense that once a model has been built for a specific semantic type (e.g. people, companies or faces) it can be used for joining any two datasets which share that semantic type.  It is also efficient because indexing uses space partitioning algorithms (such as approximate nearest neighbor) to find surface forms that are potentially joinable, thus eliminating large parts of the vector space from consideration.  Further, nearest neighbor algorithms have been applied to billions of vectors \cite{JDH17}, so the approach is practical for most datasets.  

To test the feasibility of these ideas, we used Wikidata as ground truth to build models for datasets with 1.1M people's names (about 200K identities) and 130K company names (70K identities).  The problem of mapping vectors for the same entity closer in vector space than vectors for other entities is known in the literature as deep metric learning.  Deep metric learning is known to be a difficult problem as studied in the space of face recognition and person-re-identification \cite{DBLP:conf/cvpr/SchroffKP15}, \cite{8445716}, \cite{DBLP:journals/corr/abs-1802-03170}.  As a result, there is a significant amount of research on two aspects of training these networks: (a) how to choose samples for efficient learning, and (b) what constitutes a good loss function.  In building models for entity names, we had to adapt these techniques for triplet selection and loss functions because matching entity names has different characteristics, as we describe below.

As in face recognition, our system for metric learning is built by training a so-called triplet based `siamese triplet' network to learn to produce a small distance estimate for two surface forms for the same entity (between an arbitrarily chosen \textit{anchor} and \textit{positive}), and a large distance estimate for surface forms of different entities (an \textit{anchor} and a \textit{negative}).  A key problem in effective training of such networks is the problem of how to select negative pairs for training, because one cannot exhaustively show all negative pairs to the network.  In prior work for instance, this problem is solved by so-called `triplet mining' where so-called `semi-hard' negatives are gleaned after an all-pairs comparison of input vectors in a batch, e.g., \cite{DBLP:conf/cvpr/SchroffKP15}.  `Semi-hard' negatives are negatives that are further away from the anchor than positives, but not by a sufficient margin.  The idea of focusing on semi-hard negatives is to avoid mining noisy regions of the embedding space.  However, for matching across entity names, most positive forms for the entity names are actually substantially further away from the anchor than are negatives, as we show empirically in this paper.  In other words, `hard negatives' dominate the space, and are the norm.  Our approach to the problem of triplet mining was therefore to use an approximate nearest neighbors algorithm to choose negatives for training that were in the nearest neighbor set, which means that most of the time, our negatives are `hard negatives', where the negative is closer to the anchor than positives.  This approach has three key benefits.  First, it lays out the entities based on their input vectors, and thus allows an efficient gathering of all nearest neighbors without a quadratic comparison of vectors in a batch to determine suitable negatives.  Second, it provides a baseline against which one can objectively measure the effects of training.  Third, it examines whether focusing on hard negatives is really detrimental to deep metric learning, at least for learning to map entity names in vector space.  As we show empirically in this paper, this technique is better for building models that are suitable for use in joins than semi-hard triplet mining when the dataset is dominated by hard negatives.  For easier datasets, the nearest neighbors approach was just as good as semi-hard training.

We also investigated the effect of using multiple local loss functions which have been proposed in the literature for improving deep metric learning, e.g., the triplet loss \cite{DBLP:conf/cvpr/SchroffKP15}, improved loss \cite{Zhang:2016:DML:3088616.3088665}, and angular loss \cite{DBLP:journals/corr/abs-1708-01682} functions.  For the problem of building models for datasets dominated by hard negatives, we found that an adaptation to the triplet loss function proposed in \cite{DBLP:conf/cvpr/SchroffKP15} was better than three other functions that have been used in the literature. All code\footnote{\url{https://github.com/yehudagale/fuzzyjoiner}} and models\footnote{\url{https://drive.google.com/open?id=1zivCTGkq2_AkfjGLHMnlehzTmYUwcQ9e}} for this paper are available.

\section{Related Work}
Extensions in data management systems for handling joins typically use string similarity algorithms such as edit-distance, Jaro-Winkler and TF-IDF; e.g., \cite{Cohen2003}.  String matching algorithms often do not work for merging different forms of the same entity because valid transformations of entity names can yield very different strings.  More recently, data driven approaches have emerged as a powerful alternative for merging data.  Data driven approaches mine patterns to determine the `rules' for joining a given entity type.  One example of such an approach is illustrated in \cite{He:2015:SJS:2824032.2824036}, which determines which cell values should be joined based on whether those cell values co-occur on the same row across disparate tables in a very large corpus of data.  Another example is work by \cite{auto-join-joining-tables-leveraging-transformations} where program synthesis techniques are used to learn the right set of transformations needed to perform the entity matching operation.  Our approach for merging datasets is much more general than either approach because the mapping function generalizes the set of transformations that are allowed across surface forms of an entity, even if they cannot be directly expressed as program transforms.

The idea of building joint embeddings for merging datasets followed by nearest neighbors search has been applied recently to the problem of linking relational tuple embeddings with embeddings of other relational tuples or unstructured text \cite{Bordawekar18}, \cite{IDEL18}).  For the problem of linking tuples, each model that is learnt is specific to the database it was trained on.  Our focus is on techniques that can be used to develop general purpose embedding models for merging alternate surface forms of key entities.  Once such models are built, they can be applied to joining any two datasets that share that semantic type.

Metric learning is a well studied problem in the face recognition literature, e.g., \cite{DBLP:conf/cvpr/SchroffKP15}, with a rich literature in triplet mining techniques.  The closest approach to ours is the use of nearest neighbors algorithms for semi-hard triplet mining \cite{DBLP:journals/corr/KumarHC0D17}.  For semi-hard triplet mining, one cannot look at fixed neighborhood sizes in building triplets.  If all the positives are further away from the anchor than the negatives in a given neighborhood size of $k$, it means that $k$ needs to be expanded until a neighborhood size is found that has the right characteristics.  Our approach in including `hard negatives' means we can use a fixed $k$ to generate samples.  An additional benefit is that at least for certain types of datasets, we show that metric learning with hard negatives is more effective than semi-hard mining.

The study of loss function effectiveness in metric learning is also a rich literature, with two basic types of loss functions that have been proposed: (a) local loss functions such as triplet loss \cite{DBLP:conf/cvpr/SchroffKP15}, angular loss \cite{Zhang:2016:DML:3088616.3088665} and improved loss \cite{DBLP:journals/corr/abs-1708-01682}, and (b) loss functions that operate on a more global level across a batch of training examples \cite{NIPS2016_6200}, \cite{DBLP:conf/cvpr/SongXJS16}, \cite{songCVPR17}.   Since our triplet selection is global, rather than batch based, we did not see the value of using global loss functions.

\section{Network Architecture}
\label{architecture}
\begin{figure}
\includegraphics[width=1.0\linewidth]{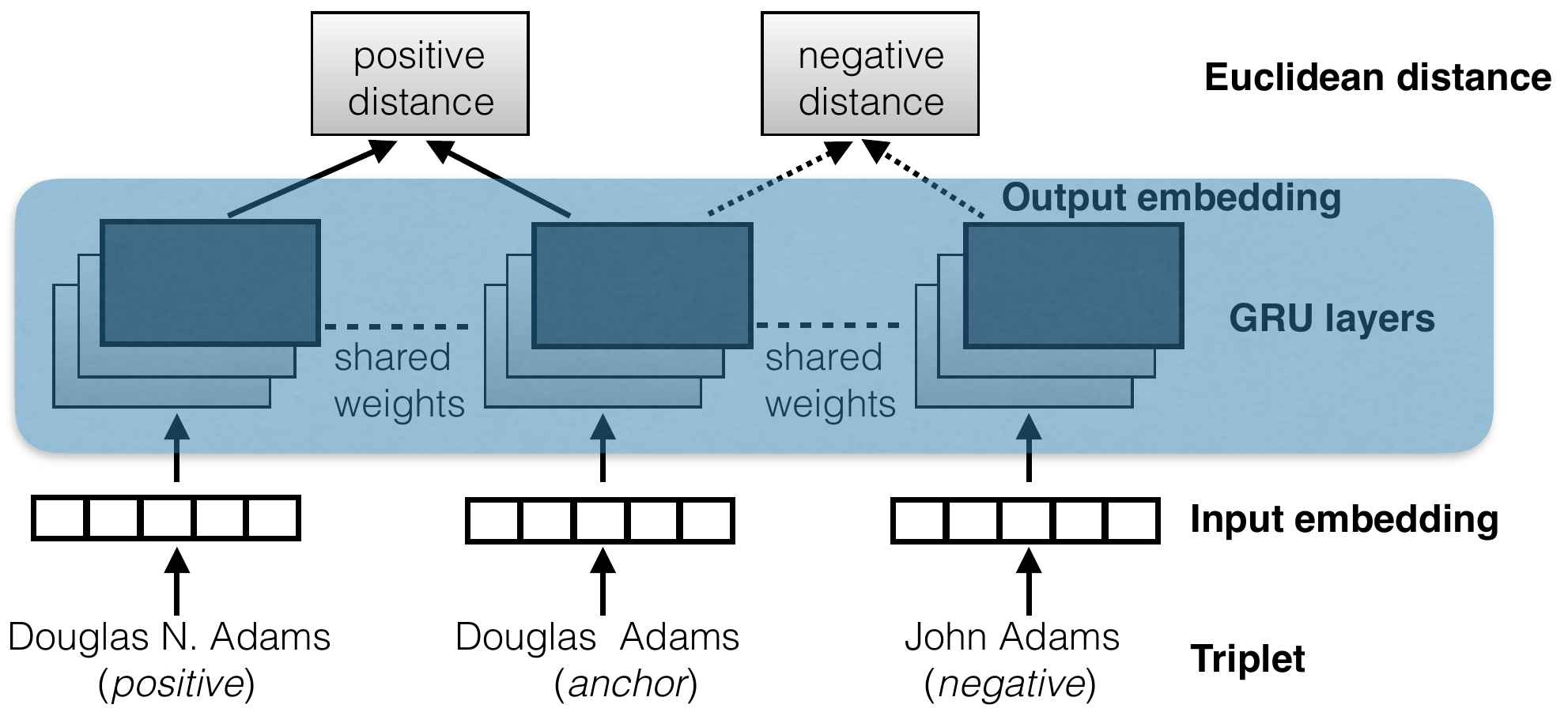}
\caption{Architecture of the siamese network}
\label{siamese_nets}
\end{figure}

Figure~\ref{siamese_nets} illustrates the siamese network architecture \cite{DBLP:conf/cvpr/SchroffKP15} we implemented to build the models using Keras and Tensorflow.  As stated earlier, input is initially triples of a name, i.e., an \textit{anchor}, a \textit{positive} and a \textit{negative}.  During the tokenization process, we kept punctuation such as `-', `,', and `.' in the name because they are important signals in processing a name.  For the network, input vectors are computed for each entity assuming a maximum name length of 10 for each entity.  A 100 dimensional character embedding was computed for each token in the entity using pretrained character embeddings \cite{hashimoto-jmt:2017:EMNLP2017}, which resulted in a 100 x 10 character encoding for each name used in a triplet.  We used character rather than word embeddings primarily because many names were missing from word based vector embeddings.

These three input character embedding vectors were fed to three identical networks that share the same weights.  Weight sharing ensures that the networks learn the same mapping function.  In our implementation, each of the three networks had 4 stacked layers of 128 unit Gated Recurrent Units (or GRUs) to capture the sequential nature of the input.  GRUs are a type of recurrent network \cite{cho-al-emnlp14} where each hidden unit updates its weights at a specific step in the sequence $t$ based on the current input $x_t$ and the value of the hidden unit from the prior step $h_{t-1}$.  For name and textual data, positional information is critical, so we used GRUs instead of the convolutional neural networks (CNNs) that have been traditionally used in metric learning for face and image recognition. 





The output of the last layer shown in the Figure~\ref{siamese_nets} as a dark layer is the vector embedding for the inputs.  These are fed to two layers which compute a euclidean distance between the \textit{anchor} and the \textit{positive} (\textit{positive distance}), and the \textit{anchor} and the \textit{negative} (\textit{negative distance}).  Conceptually, there are two objectives in metric learning, one to minimize \textit{positive distances}, and the other to maximize \textit{negative distances}.  As described below, this dual objective can be achieved by different loss functions.  We restrict ourselves to a discussion of the some of the more popular loss functions that are local in nature (i.e., they only look at a single triple).  

\section{Loss functions}
\label{loss_functions}

Let $\mathbf{x}$ represent an embedding for an entity name, and $\bf{x_{a}}$, $\bf{x_{p}}$, $\bf{x_{n}}$ reflect the vector embeddings of the \textit{anchor}, \textit{positive} and \textit{negative}, respectively.  We investigated four different loss functions, three of which have been used in prior face recognition literature to explore their effectiveness for the entity metric learning problem.

\begin{figure}[htb]
    \centering
    \begin{subfigure}[t]{0.31\linewidth}
        \centering
        \includegraphics[width=\linewidth]{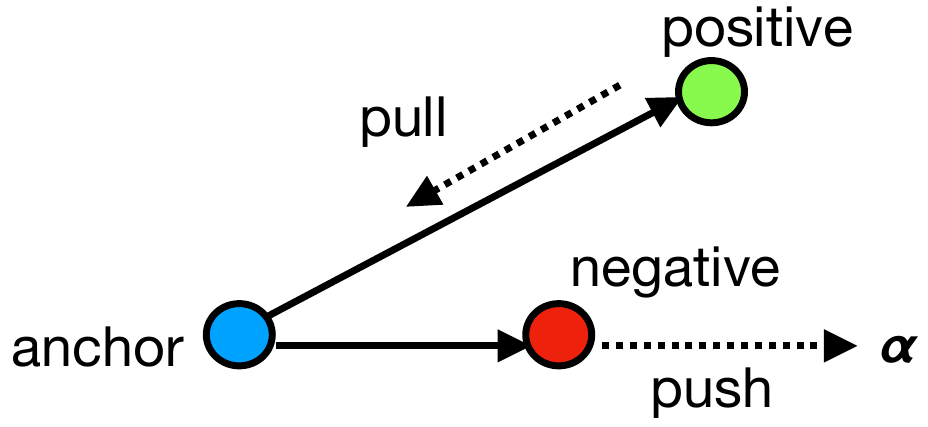}
        \caption{Triplet loss}
        \label{schroff_loss}
    \end{subfigure}%
    ~ 
    \begin{subfigure}[t]{0.31\linewidth}
        \centering 
        \includegraphics[width=\linewidth]{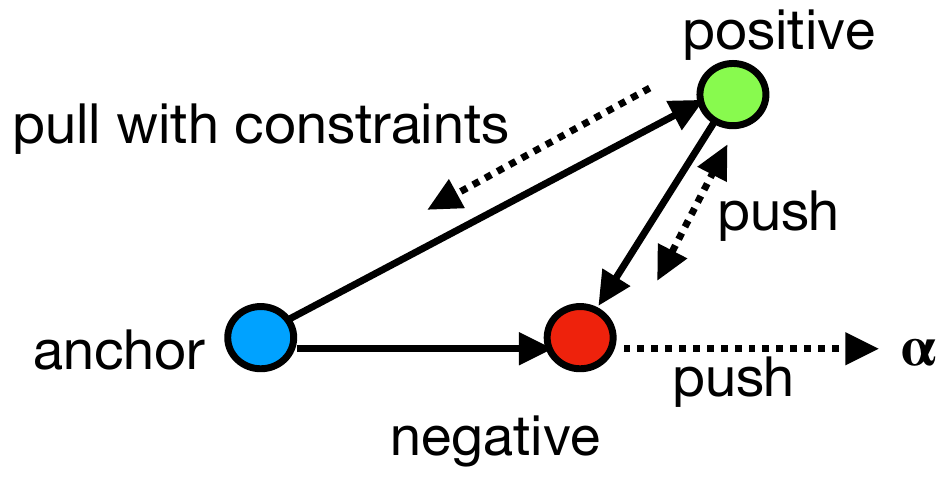}
        \caption{Modified loss}
        \label{modified_loss}
    \end{subfigure}
    ~ 
    \begin{subfigure}[t]{0.31\linewidth}
        \centering 
        \includegraphics[width=\linewidth]{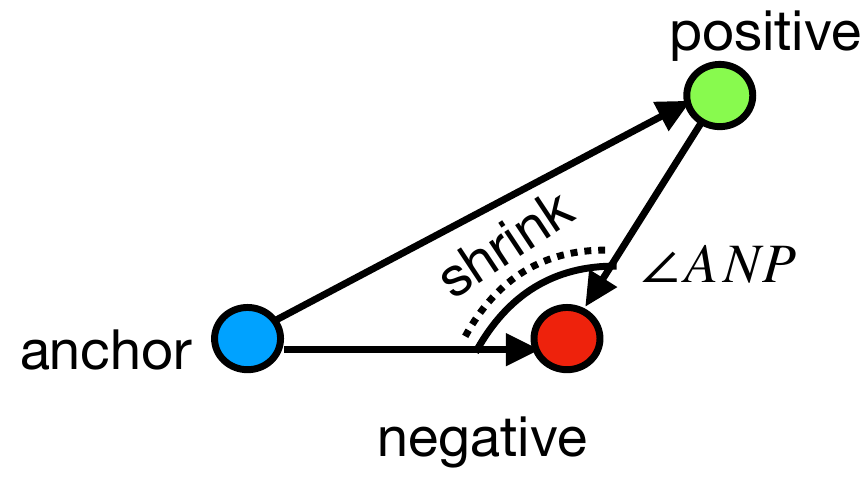}
        \caption{Angular loss}
        \label{angular_loss}
    \end{subfigure}
    \caption{Loss functions}
\end{figure}

\subsection{Triplet loss}

For face recognition, Schroff et al. \cite{DBLP:conf/cvpr/SchroffKP15}
propose a triplet loss function where the \textit{positive distances}
for each triplet $i$ in the set of $N$ triplets is separated from
\textit{negative distances} by a margin of $\alpha$, as shown in
Figure~\ref{schroff_loss} with the arrow pushing toward $\alpha$.  For
each of $N$ triples, $l_{triple}$ reflects the loss for a given triple
as follows: 
\begin{equation}
  l_{triple} =  \left[\|\bf{x_a} - \bf{x_p}\|^2 - \|\bf{x_a} -\bf{x_n}\|^2 + \alpha \right]_+
\label{schroff}
\end{equation}
where
\begin{equation}
 \left[.\right]_{+} = max(0, .)
\end{equation}
and the loss function that is minimized across all $N$ triples is given by
\begin{equation}
 L = \sum_{i}^{N} l_{triple}
\end{equation}
Note that in this formulation, it is assumed that embedding is normalized so $\|\bf{x} \| = 1$ because this normalization is robust across variations in illumination and contrast.  The value of $\alpha$ in the original work is a hyper-parameter that \cite{DBLP:conf/cvpr/SchroffKP15} was set to 0.2.

\subsection{Improved Loss}

 An improvement over the triplet loss function is proposed by
 \cite{DBLP:conf/cvpr/SchroffKP15} for the recognition of faces in
 videos \cite{Zhang:2016:DML:3088616.3088665}.  Conceptually, this
 function that we refer to as `improved loss' in the paper considers
 the distances from the \textit{positive} to the \textit{negative}, and tries to push that
 difference toward $\alpha$ as well as the distance from
 \textit{anchor} to the \textit{negative}.  We show this in
 Figure~\ref{modified_loss} with two push arrows.  In addition, it corrects the fact the  original triplet loss function has no constraints on how close the positive distance should be.  For instance, it is possible for the \textit{anchor} and \textit{positive} form a large cluster with a large intra-class distance. The equations that achieve these constraints are described below.  Equation~\ref{psi} tries to reduce intra-class distance by ensuring it is less than $\hat{\alpha}$.  Equation~\ref{phi} tries to maximize inter-class distance by ensuring that the distance from the \textit{anchor} and \textit{positive} to the negative are both taken into account.  Equation~\ref{improved_loss} balances inter-class constraints with intra-class constraints with the parameter $\lambda$. 

\begin{equation}
  \psi_{triple} = \|\bf{x_{a}} - \bf{x_{p}}\|^2 - \hat{\alpha}
\label{psi}
\end{equation}
\begin{equation}
  \phi_{triple} = \|\bf{x_{a}} - \bf{x_{p}}\|^2 - (\|\bf{x_{a}}- \bf{x_{n}}\|^2 + \|\bf{x_{p}} - \bf{x_{n}}\|^2) / 2  + \alpha
\label{phi}
\end{equation}
\begin{equation}
  l_{triple} = max(0, \phi) + \lambda * max(0, \psi)
\label{improved_loss}
\end{equation}
The parameter $\hat{\alpha}$ for equation~\ref{psi} is set to 0.1, and $\lambda$ is set to .02 in equation~\ref{improved_loss}, and $\alpha$ was set to 1 in the paper \cite{Zhang:2016:DML:3088616.3088665}.  As in Schroff et al. \cite{DBLP:conf/cvpr/SchroffKP15}, the embeddings are normalized to 1 although the actual paper does not make it clear.

\subsection{Angular Loss}

Wang et al. \cite{DBLP:journals/corr/abs-1708-01682} define a novel
angular loss function which is not based on pairwise distances, but
rather is based on the angles of the triangle formed by the
\textit{anchor}, \textit{positive} and the \textit{negative} triplet.
Conceptually, they point out that since the \textit{anchor} and the
\textit{positive} pairs belong to the same class, the angle formed by
the \textit{anchor}, \textit{negative}, and \textit{positive} elements
should be as small as possible within that triangle.  Their loss
function is an attempt to minimize this angle, as defined in the
equations below.  The rough idea is that moving the \textit{positive} nearer
and the \textit{negative} further away each reduce the angle, which we
illustrate in Figure~\ref{angular_loss} with the angle to shrink.
\begin{equation}
\bf{x_{c}} = (\bf{x_{a}} + \bf{x_{p}}) / 2
\end{equation}
\begin{equation}
l_{triplet} = \left[[\bf{x_{a}} - \bf{x_{p}}\|^2 - 4 \tan^2 \alpha \|\bf{x_{n}} - \bf{x_{c}}\|^2 \right])_{+}
\end{equation}

\subsection{Adapted Loss}

The inspiration for adapting the original triplet loss is to separate the effect of the positive
and negative distances as much as possible.  Rather than subtracting
the negative distance from the positive one, we want to negate the
negative distance and then add the two.  To approximate this, we
subtract the negative distance from a margin, and use 0 instead if
that difference is negative.  We then combine the squared distances as
usual, as shown in Figure~\ref{adapted}.  Note that we did not normalize $\|\bf{x} \| = 1$ because learning was worse with normalization.

\begin{equation}
  l_{triple} =  \|\bf{x_a} - \bf{x_p}\|^2 + \left[ \alpha - \|\bf{x_a} -\bf{x_n}\| \right]_+^2
\label{adapted}
\end{equation}

\section{Triplet Selection}
As discussed earlier, deep metric learning for joins is a difficult problem for neural networks to learn because it requires that the discrimination of each anchor from all the other hard negatives.  We describe a popular approach batch based approach from the face recognition literature first to contrast it with our mechanism for triplet selection.

\subsection{Batch Based Triple Selection}
\begin{figure}
\includegraphics[width=1.0\linewidth]{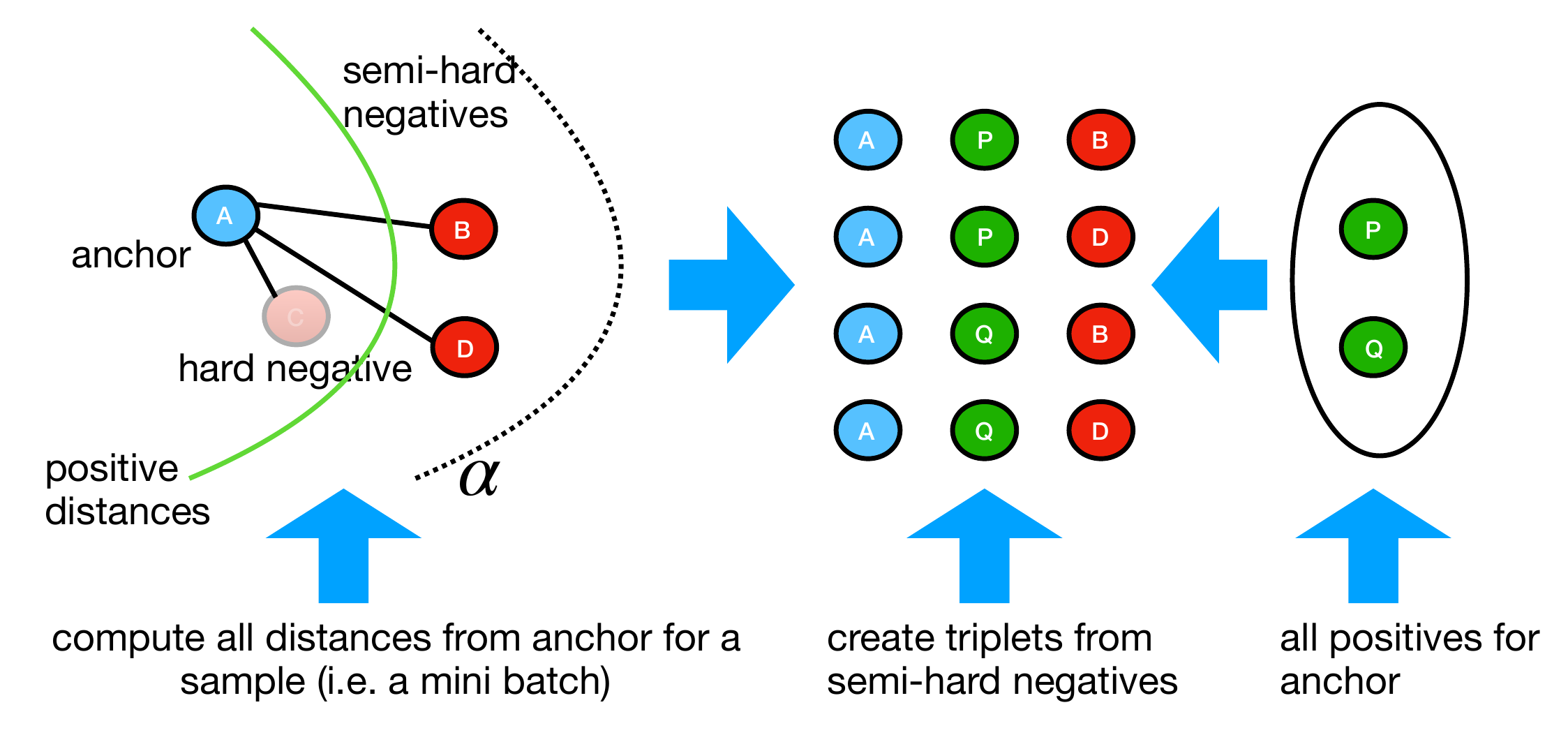}
\caption{Batch based triplet selection}
\label{triplet_selection}
\end{figure}

In Schroff et al.'s work~\cite{DBLP:conf/cvpr/SchroffKP15} they described a mini-batch based triplet selection mechanism for training that has dominated the literature.  Conceptually, sampling the right triplets for fast network learning requires sampling a set of \textit{hard positives} and a set of \textit{hard negatives}, where a hard positive is defined as $argmax^i \| x_{a} - x_{p}^i \|^2$, where $i$ ranges over all $x_p$ and a hard negative is defined as $argmin^i \| x_{a} - x{_n}^i \|^2$, where $i$ ranges over all $x_n$. 

However, it is infeasible to compute these values for the entire dataset.  Calculating hard positives is easy because the number of \textit{positives} is small normally.  Calculating hard negatives is not possible for all but small datasets.  As a result, the triplets can be generated by a mechanism illustrated in Figure~\ref{triplet_selection}.  Instead of focusing on finding \textit{hard positives}, they instead pair every possible positive in the sample shown in the right panel in the figure with selected negatives, since the set of positives is usually quite small.  Furthermore, for negative examples, they try to select \textit{semi-hard negatives} instead of \textit{hard negatives}, where a \textit{semi-hard negative} has the property that $\|x_a - x_p \|^2 < \|x_a - x_n \|^2$ but by a margin smaller than $\alpha$, as shown in figure ~\ref{triplet_selection}.  


\subsection{Metric Based Triplet Selection}
For the problem of joins, we ideally want the anchor and all of its positives to be clustered closest together and separated from the nearest negatives as clearly as possible.  Approximate nearest neighbor (ANN) indexes are highly efficient methods for selecting the top-K neighbors of a given vector by Euclidean distance, cosine similarity or other distance metrics.  They are based on space partitioning algorithms, such as \textit{k-d trees}, where the vector space is iteratively bisected into two disjoint regions.  The average complexity to query the vectors of a neighbor is $O(log N)$ where N is the number of vectors in the dataset.  Implementations exist now for fast, memory-efficient ANN indexes that scale up to a billion vectors \cite{JDH17} using techniques to compress vectors in memory efficiently.  In our work, we used the Annoy ANN implementation\footnote{\url{https://github.com/spotify/annoy}} in our work which is based on the refinements to \textit{k-d trees} ideas described in \cite{ann_paper}.

\begin{figure}
\includegraphics[width=1.0\linewidth]{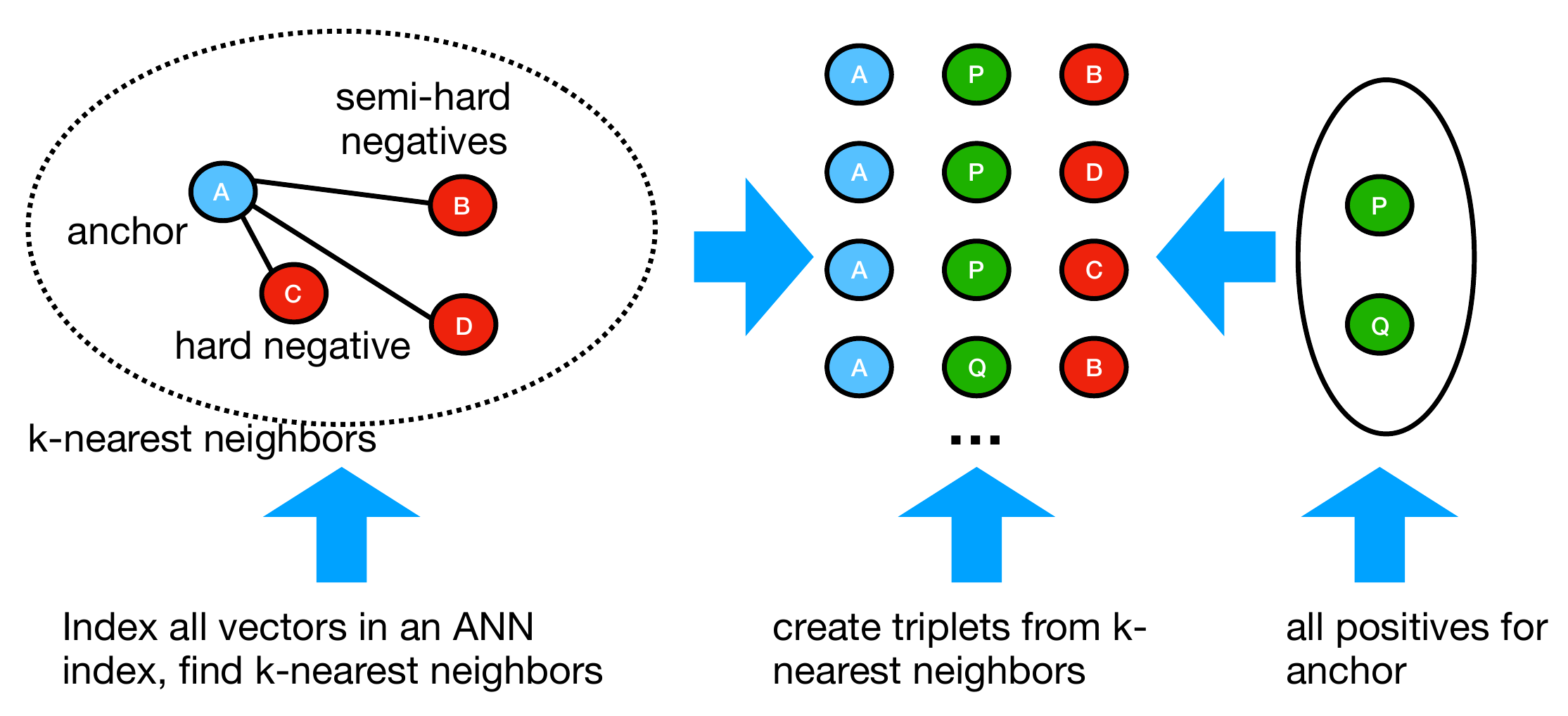}
\caption{ANN based triplet selection}
\label{ANN_selection}
\end{figure}

Assuming one has the entire dataset indexed in an ANN index, the problem of triplet selection can be simplified by asking the ANN index for the top k-nearest neighbors of an \textit{anchor}, where k is given by the number of triplets that one desires to generate for each anchor.  As in earlier work, selection of \textit{hard positives} is not relevant because all positive data should be used to teach the network the right function.  Selection of negatives is the set of all nearest neighbors that are \textit{negatives}.  There is no explicit attempt to filter out \textit{hard negatives} in the approach.  The overall idea is that the set of negatives that appear in the nearest neighbor set at input are in fact the most important elements for the network to learn to discriminate from positives for a join.  Focusing on these elements, regardless of whether they are \textit{hard} or \textit{semi-hard} should lead to better discrimination for joins.  An ANN-based strategy also provides an important baseline to assess what if any learning was performed by the neural network in mapping input vectors to a different space.

\section{Applying deep learning models to joins}
\label{join_system}


Assuming we have deep learning models that are trained to produce the
right distance estimates for alternate surface forms for an entity,
the models can be used for a join as follows.  For each cell value in
the two columns to be joined, obtain vector embeddings from the last
layer of the network.  Note that although the siamese network has
three separate networks, each network is in fact identical to the
other two networks because they share weights.  For the left column
cell values, vector embeddings are inserted into an approximate
nearest neighbors index.  For each cell value in the right column,
vector embeddings are used as `query vectors' to query the approximate
nearest neighbors index.  In our context, merging the datasets would
involve joining the top $k$ rows in the left table that are `closest'
in distance to each cell value in the right table.  Note that the
choice of $k$ clearly has a direct effect on the tradeoff between
precision and recall, but for most practical uses of join, $k$ is
usually very small (typically 1).  This has implications on what
metrics we can use to evaluate joins, as we describe in our evaluation.

\section{Benchmarks}
\label{datasets}
Our benchmarks were derived from Wikidata.  Specifically, we used the \textit{also known as} property from Wikidata to get alternate forms for the same entity name for people as well as companies.  For company names, we augmented the names and surface forms in Wikidata with data from the SEC\footnote{\url{https://www.sec.gov/dera/data/financial-statement-and-notes-data-set.html}}, which has former and more recent names for companies.  There were 213,106 names for people from the specific dump we extracted, and 70,946 names for companies.  The extracted files and the cleansing code are available on our repository. The extracted files however contained significant noise that we cleaned up programmatically.  We describe the cleansing rules for people and for companies separately because they were somewhat different.  In the case of people's names, we also augmented the data so the system could learn some common rules that define variants of a person's name.  This was not possible for company names.

\subsection{Cleansing people's names}
Wikidata has a number of historical figures which are not really names of people (e.g. Queen Elizabeth, Pope Leo).  If we detected a title in the name referring to royalty or qualifiers or Roman numerals which strongly indicated royalty, we dropped the name.  We also removed punctuations such as '...', and anything that was placed in parenthesese because they were not part of the name (e.g. a qualification such as the son of Jacob might appear in parentheses after a name).  Although we got the extract for English, there were frequently names in Chinese, Korean, Cyrillic, etc.  We removed these and restricted ourselves to names in ISO-8859-1 unicode.  All punctuations such as ',', '-', '.' etc were retained for names because they are strong indicators of how a name needs to be parsed.  People in wikidata have the main name for the entity, with aliases for the person specified in a different property.  We made sure that every alternate form had at least one name part in common with the main name to rule out 'nicknames' (e.g. `Father of the Nation' for George Washington).  We also dropped cases where the last name of a person was different (usually because a woman's name changed after marriage).      

\subsection{Cleansing company names}
As with people's names, we removed any text in parentheses because it usually was a qualification (e.g., IBM (company)).  We restricted ourselves to unicode ISO-8859-1.  The SEC data had a lot of strange company names that could be described with the regex pattern T[0-9]+ or [0-9]+, and we dropped these.  We tried to ensure we included names that shared some subset of characters with the main name, ensuring we would not drop acronyms when possible.  The check for acronyms tested if any of the initial letters of each name part occurred in the name.

\subsection{Augmenting people's names}
In many cases, we had no alternate forms for a person's name even if we did have their main name.  We augmented the data with the following rules.  If the main name for the person in Wikidata had two parts, we created new source forms as follows: (a) Last Name , First Name, (b) First Name Initial . Last Name, and (c) Last Name , First Name Initial.

If the main name in Wikidata had three parts, we created these additional source forms in addition to the ones listed above: (a) First Name Middle Name Initial Last Name (b) Last Name , First Name Middle Name (c) Last Name , First Name Middle Initial .

After cleansing, if a name had no alternate surface forms, they were dropped.  This resulted in 40,555 company names with an average of 2.2 names each, and 195,422 people's names with an average of 4.69 names each.  Using the triplet selection algorithm we created a set of 10.9 million and 1.04 million triplets for people and companies at training.  The data was then split with a 60-20-20 ratio to provide training, validation and test data respectively.  Each run was conducted with a different random split to ensure generalization of results.

\subsection{Dataset characteristics}
\begin{figure}[htb]
    \centering
    \begin{subfigure}[t]{0.23\textwidth}
        \centering
        \includegraphics[width=.95\linewidth]{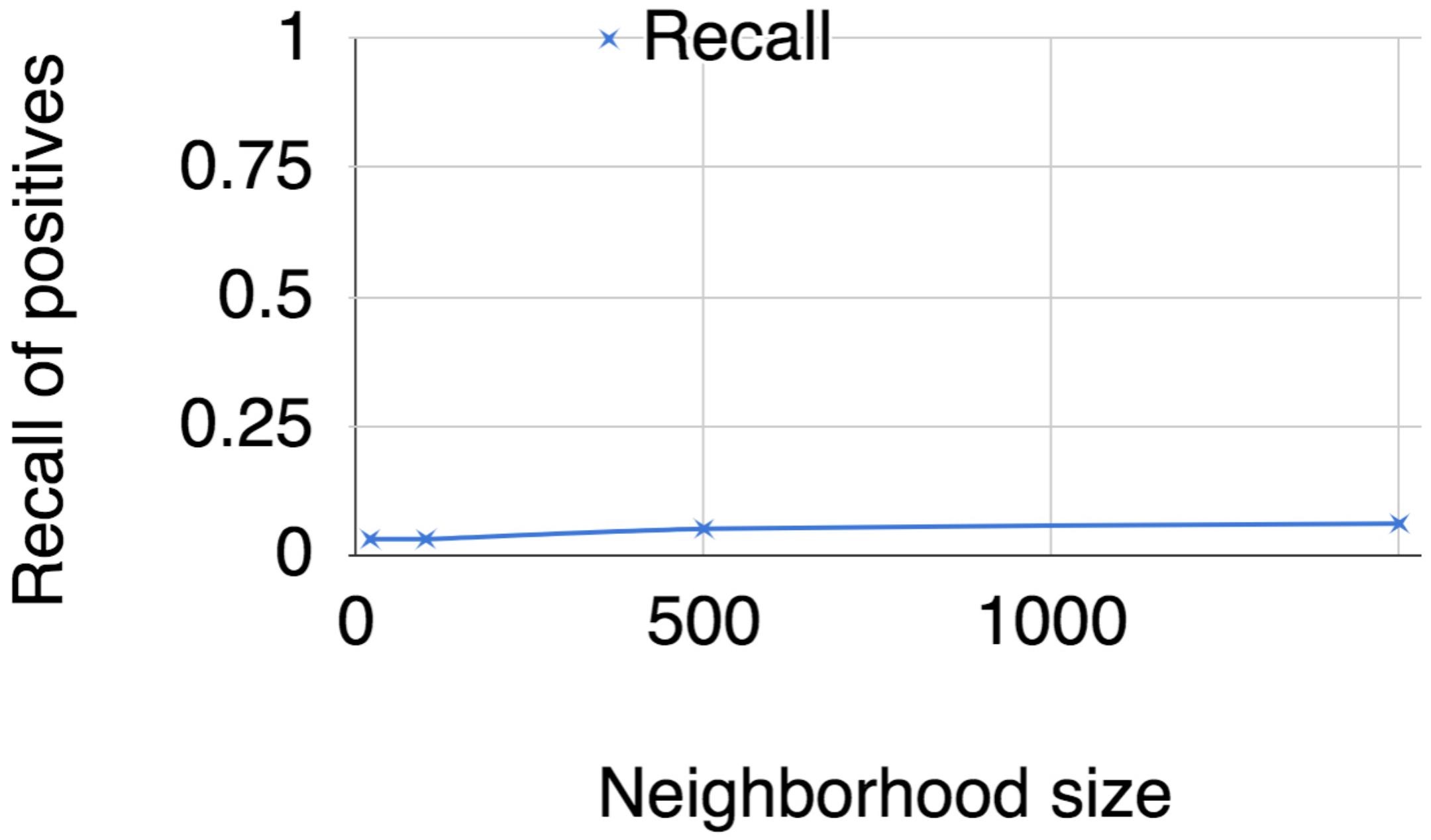}
        \caption{Recall rate as a function of neighborhood size}
        \label{people_recall}
    \end{subfigure}%
    ~ 
    \begin{subfigure}[t]{0.23\textwidth}
        \centering 
        \includegraphics[width=.95\linewidth]{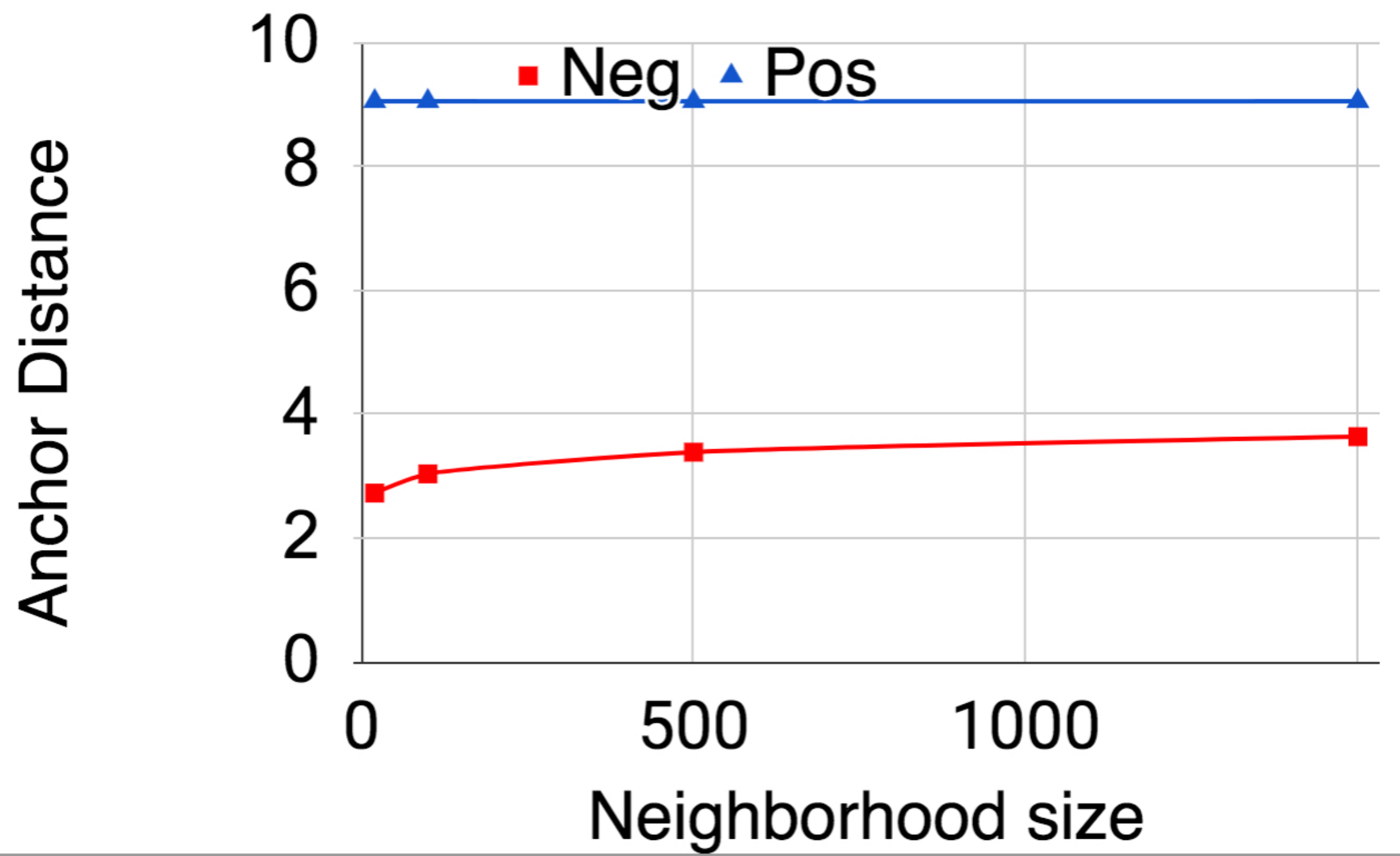}
        \caption{Mean negative distance as a function of neighborhood size}
        \label{modified_loss}
    \end{subfigure}
    \label{people_distances}
    \caption{Characteristics of people data}
\label{people_characteristics}
\end{figure}

\begin{figure}[htb]
    \centering
    \begin{subfigure}[t]{0.23\textwidth}
        \centering
        \includegraphics[width=.99\linewidth]{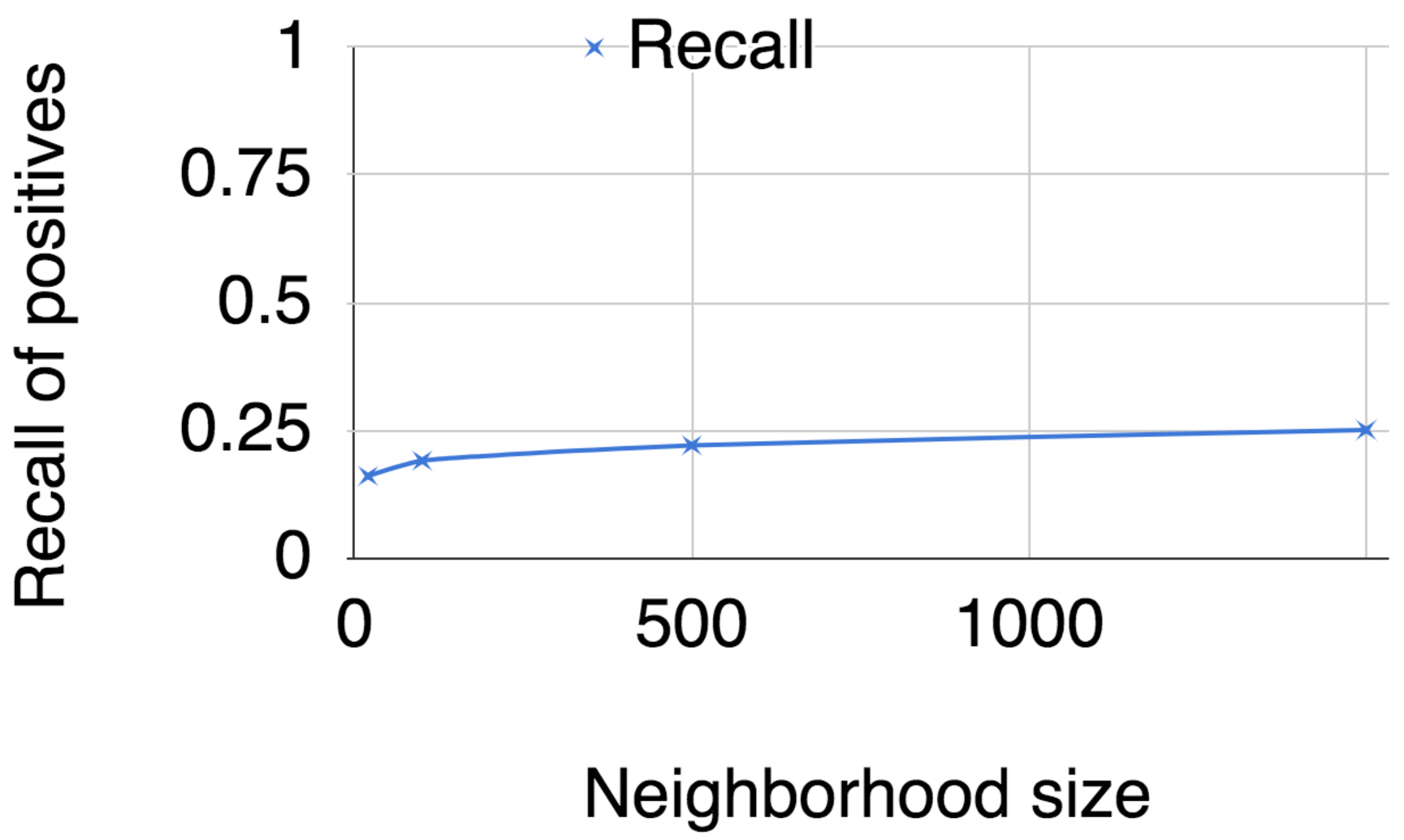}
        \caption{Recall rate as a function of neighborhood size}
        \label{company_recall}
    \end{subfigure}%
    ~
    \begin{subfigure}[t]{0.23\textwidth}
        \centering
        \includegraphics[width=.99\linewidth]{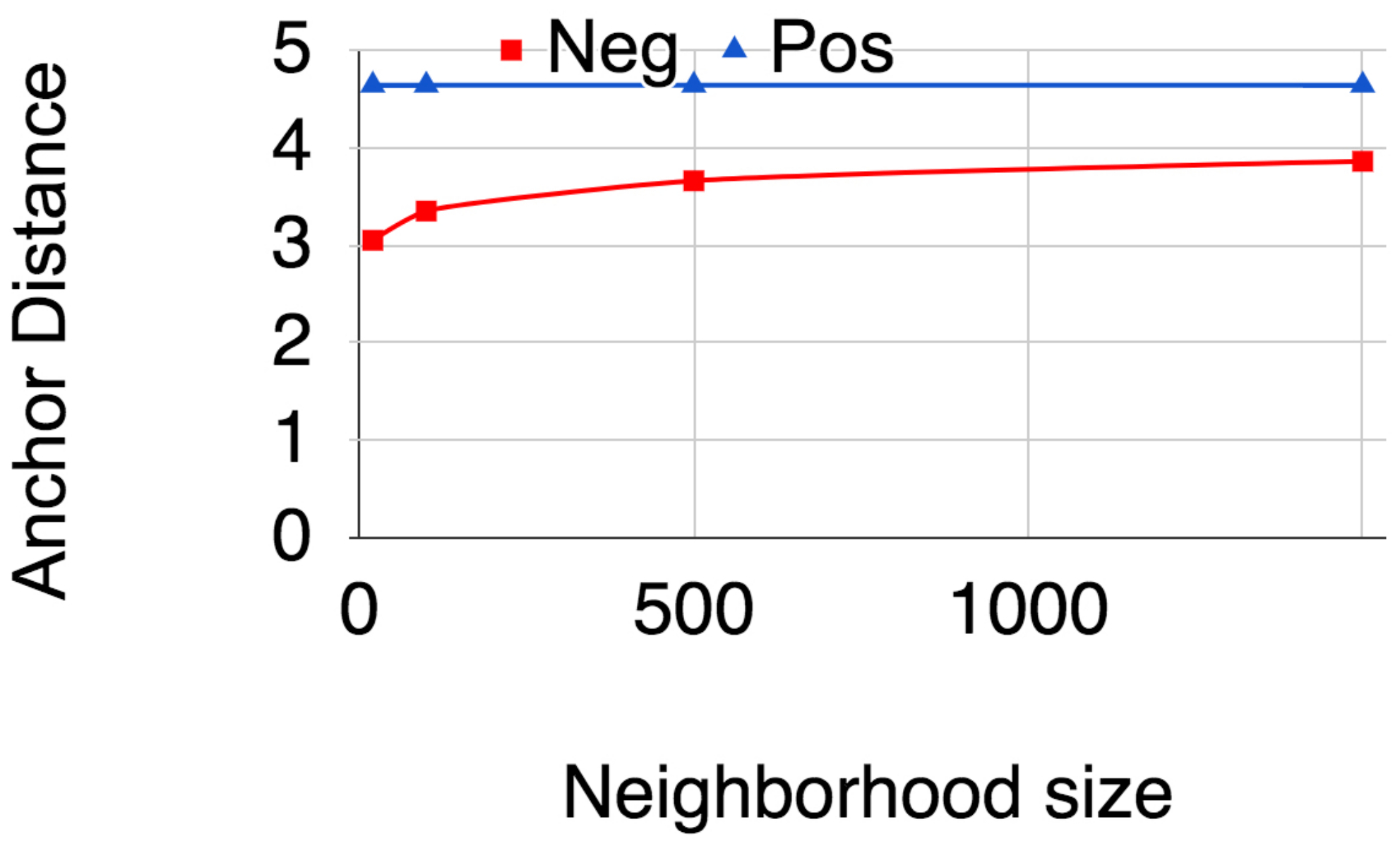}
        \caption{Mean negative distance from anchor as a function of neighborhood size}
        \label{modified_loss}
    \end{subfigure}
    \label{company_distances}
    \caption{Characteristics of company data}
\label{company_characteristics}
\end{figure}

As a baseline, we measured how the anchors, positives and negatives were laid out in vector space based on character embeddings alone.  This gives us a measure of how difficult the problem is for the neural network to learn.  We indexed all the vector embeddings (regardless of test or train) into a nearest neighbors algorithm using the Spotify ANNOY\footnote{\url{https://github.com/spotify/annoy}} implementation.  We then queried the index for the $k$ nearest neighbors of this set, varying $k$ so it was either 20, 100, 500, or 1500 neighbors.  Our primary interest was in recall rates of positives prior to any training, to establish the baseline prior to training.  We also measured the nature of hard negatives as we varied the neighborhoods; i.e., what is the mean distance of negatives from the anchor while we increased neighborhoods, compared to the positive distances.  Figure \ref{people_characteristics} shows the results for the people data.  First, recall rate of positives in the nearest neighbor set was very low at 3\%, and it increased only to 6\% at a neighborhood size of 1500.  The difficulty of the problem for reconciling people's names is further highlighted by the distance data.  The mean distance of positives from anchor was 9.05, with a standard deviation of 3.08.  The mean distances of negatives from anchor was 2.73, with a standard deviation of 0.99 for $k$ of 20.  However even for $k$ of 1500, the mean negative distance was 3.64, well below the mean positive distance of 9.05.  The company data show a similar trend, although companies seems to be an easier problem than reconciling people data, as shown in Figure \ref{company_characteristics}.  Recall rate for companies starts at 16\% for a neighborhood size of 20, and is up to 25\% for a neighborhood size of 1500.  Mean negative distance is at 3.05 compared to 4.64 at a neighborhood size of 20, with a standard deviation of 1.44.  At a neighborhood size of 1500, mean negative sizes are slightly higher at 3.86. 

\section{Experiments}
\label{results}
In building the models, we employed early stopping using the usual metric of accuracy of the validation data, and we performed hyper-parameter tuning using grid search varying margins for the adapted loss function from 1-20.  Accuracy was defined as the percentage of validation triples where positive distances from anchor were less than negative distances from anchor.  For all our runs, test accuracy as measured by this metric ranged in the 97-99\% range for the people dataset and 92-94\% companies dataset for all losses except angular loss which was poor throughout.  Table \ref{precision_recall} shows the results for the people and company datasets, run with a fixed $k$ of 20 neighbors.  Because we compared across different losses, the results are categorized by each loss function for each dataset we tested.  For all the results reported here, we ran multiple runs because of the stochastic nature of neural network models; the results here are means across two runs. 

We report multiple metrics to measure the effectiveness of training, some of which are not standard because of the experimental setup, so we define them below:
\begin{itemize}
\item \textbf{Recall}.  Recall is measured by the percentage of positives in the nearest neighbor set of each anchor.
\item \textbf{Precision@1}.  Precision@1 is measured by the percentage of anchors with the very nearest neighbor being a positive.  As pointed out earlier, this is an important metric for assessing join performance in a majority of cases.
\item \textbf{Precision}.  Precision is measured by the fractions of all positives for each anchor that were closer to that anchor than was any negative.
\end{itemize}
 
\begin{table}[ht]
\caption{Precision and Recall by loss functions}
\label{precision_recall}
\begin{center}
\begin{tabular}{|l|l|r|r|r|}
\hline
Entities & Loss & Recall & \multicolumn{2}{|c|}{Precision} \\
 & & & @1 & All \\
\hline
\multirow{4}{*}{People} & Adapted & .81 & .81 & .63 \\
\cline{2-5}
& Triplet & .74 & .84 & .55 \\
\cline{2-5}
& Improved & .71 & .52 & .45 \\
\cline{2-5}
& Angular & .04 & .09 & .02 \\
\hline
\multirow{4}{*}{Corps} & Adapted & .74 & .73 & .66 \\
\cline{2-5}
& Triplet & .74 & .75 & .67 \\
\cline{2-5}
& Improved & .74 & .72 & .65 \\
\cline{2-5}
& Angular & .26 & .32 & .22 \\
\hline
\end{tabular} 
\end{center}
\end{table}

\begin{table}[ht]
\caption{Distance estimates after training}
\label{distance}
\begin{center}
\begin{tabular}{|l|l|r|r|r|r|}
\hline
Entities & Loss & \multicolumn{2}{|c|}{Pos} & \multicolumn{2}{|c|}{Neg} \\
& & Mean & Std & Mean & Std \\
\hline
\multirow{4}{*}{People} & Adapted & 1.59 & 4.05 & 2.44 & 2.45 \\
\cline{2-6}
& Triplet & .46 & .18 & .55 & .10 \\
\cline{2-6}
& Improved & .15 & .38 & .21 & .18 \\
\cline{2-6}
& Angular & .51 & .25 & .06 & .02 \\
\hline
\multirow{4}{*}{Corps} & Adapted & 2.39 & 3.62 & 3.59 & 2.72 \\
\cline{2-6}
& Triplet & .43 & .31 & .59 & .09 \\
\cline{2-6}
& Improved & .32 & .50 & .42 & .19 \\
\cline{2-6}
& Angular & .07 & .04 & .04 & .01 \\
\hline
\end{tabular}
\end{center}
\end{table}

\subsection{Performance for Joins}
For fuzzy joins, we need both precision and recall to be high.  Without good recall, a join will potentially miss names that should be joined.  Without high precision, a join will mistakenly join many inappropriate names.  The numbers for the adapted loss function were 81\% for people and 74\% for companies, so a join could capture most similar names.  For people, the adapted loss function showed overall better performance than triplet loss (see Table \ref{precision_recall}).  Furthermore, it appears that triplet loss outperformed improved loss, which in turn is better than angular loss for learning this problem.  On the other hand, for companies the only significant difference was that angular was worse.

 Precision is a little trickier to define, but one way is to measure it is to examine how many true matches we get in the neighborhood of each anchor before seeing a single mistake, i.e. a match that should not be there.  That metric gives a picture of how many names would be correctly found, on average, by a join.  Adapted loss is at 63\% by this metric for people and 66\% for companies, so about two thirds of recalled names would be found before finding a single error.  Since every name has at least one other name for the same entity, we also measure precision at 1, which is the probability that the very nearest neighbor is a true match.  For that, the very best performance we get is 84\% for people and 75\% for companies.

\subsection{Learning Performance}

 We also assessed our learning mechanism more directly by examining how the nearest neighborhood changed from before to after training.  As can be seen from Table \ref{distance}, recall is improved greatly, with the bigger change for people, in which case it improves from 3\% to 81\%.  For companies, it is from 16\% to 74\%.  Thus training is clearly effective in moving actual names for the same entity into the nearest neighbors.  For precision, the fraction of true matches found before the first error improves from 16\% to 73\% for people and 16\% to 64\% for companies.  Precision@1 improves from 10\% to 81\% for people and from 26\% to 75\% for companies.  In both these cases, demonstrable training occurred.

\subsection{Comparison with Semi-Hard Negatives}
We hypothesized that training against hard negatives can potentially benefit on datasets with characteristics like those of our people dataset, when compared to training against semi-hard negatives.  Such datasets have large numbers of hard negatives, as suggested by Figure~\ref{people_characteristics}, which shows positive distances higher on average than negative ones.  

We therefore compared the hard negative triplet selection mechanism directly to training against semi-hard negatives.  To compute semi-hard negatives, we took, for each entity, all its positives, and found all negatives in the nearest neighborhood of that positive that were further from the entity than is the positive.  We made triplets for each such pair of positive and negative.  We thus chose the hardest semi-hard triplets: the negatives are as close to the positive as possible while still being further from the entity.  We ran experiments again for adapted loss on our people dataset, changing only the triplets used; these results are in Table~\ref{hard-semi-hard}.  The same comparison on the company dataset showed no difference, mostly because company data seems easier and seems more immune to differences in loss functions or training regimens.

\begin{table}[ht]
\caption{Precision and Recall for hard and semi-hard training}
\label{hard-semi-hard}
\begin{center}
\begin{tabular}{|l|r|r|r|}
\hline
Training & Recall & Precision@1 & Precision \\
\hline
hard & .81 & .81 & .63 \\
\hline
semi-hard & .63 & .61 & .43 \\
\hline
\end{tabular}
\end{center}
\end{table}

 The results show training on hard negatives produces consistently better results, both for precision and recall.  Hard negative training resulted in recall of 81\% of positives in the nearest set versus 63\% for semi-hard negatives.  Precision@1 is similar, with 81\% for hard negatives versus 61\% for semi-hard.  Overall precision, defined as the fraction of positives closer than any negative is 63\% for hard negatives versus 43\% for semi-hard.

\subsection{Generalization Test on Faces}
We have demonstrated that our strategy for joins could work well for textual names of people and companies, but the technique could potentially work for any kind of data for which a vector embedding can be made.  To test how well that works, we evaluated two existing models for face recognition that were trained with the same approaches defined in \cite{DBLP:conf/cvpr/SchroffKP15} on two different datasets, VGGFace2 \cite{DBLP:conf/fgr/CaoSXPZ18}, and CasiaWebFace \cite{DBLP:journals/corr/YiLLL14a}.  We took the two open sourced models\footnote{\url{https://github.com/davidsandberg/facenet}}, and extracted the output embeddings for faces from the LFW test set \cite{Huang2012a} using these embeddings.  We put these output embeddings into an ANN structure, and computed our metrics on that.  Note that for face datasets there is a greater number of variability of faces per identity with a maximum of 529 faces for a single identity.  We adjusted neighborhood length to a maximum of 20 or the number of expected neighbors in our experiments for each face.  Recall was 91\% on the VGGFace2 dataset, and 87\% for the Casia Web Face dataset.  Precision@1 was 95\% for VGGFace2 and 93\% for Casia Web faces.  Overall precision was 91\% for VGGFace2 and Casia Web faces was 87\%.  These results suggest that existing models can in fact be re-purposed for joins.  We point out that the face models seem better in terms of performance compared to our model for names but there are significant differences in the data characteristics for training.  The names data had a lot fewer positives per identity.  VGGFace2 has 362.6 faces per identity, and in CasiaWebFace, 500,000 images exist for 10,000 identities.

\section{Conclusion and Future Work}
We show that deep learning models can be used effectively for joins, and we provide these models to the community for use.  In the future, we will evaluate our work on other entity types and continue to explore refinements of both loss functions and triplet selection.

  \bibliography{paper}
  \bibliographystyle{aaai}
  \end{document}